\documentclass[10pt,twocolumn,letterpaper]{article}

\usepackage[pagenumbers]{cvpr} % To force page numbers, e.g. for an arXiv version

\usepackage{cvpr}
\usepackage{times}
\usepackage{epsfig}
\usepackage{graphicx}
\usepackage{amsmath}
\usepackage{amssymb}
\usepackage{booktabs}
\usepackage{multirow}
\usepackage{xcolor}
\usepackage{colortbl}
\usepackage{pifont}

\usepackage{algorithm}
\usepackage{algpseudocode}
\definecolor{cvprblue}{rgb}{0.21,0.49,0.74}
\usepackage[pagebackref,breaklinks,colorlinks,allcolors=cvprblue]{hyperref}

\def\confName{CVPR}
\def\confYear{2027}

\title{RDGSplat: Render-Dedicated Geometry for Novel View Synthesis}

\author{Zhijie Zheng, Xinhao Xiang, Jiawei Zhang\\
University of California, Davis\\
CA 95616, USA\\
{\tt\small \{zhjzheng, xhxiang, jiwzhang\}@ucdavis.edu}
}

\begin{document}

\twocolumn[{
  \maketitle
  \begin{center}
    \includegraphics[width=1.0\textwidth]{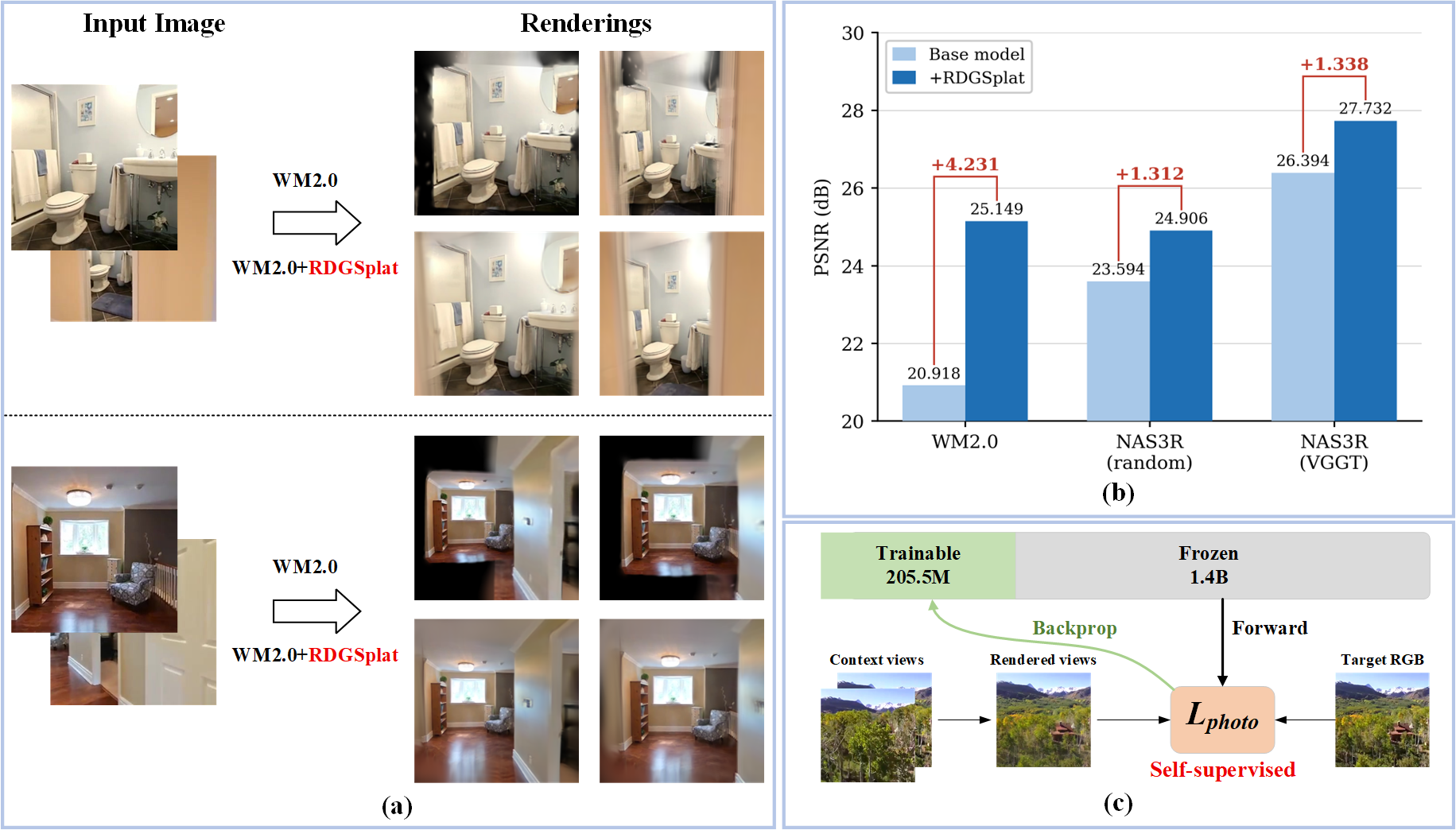}
    \captionof{figure}{\textbf{Novel view synthesis with all parameters of the original backbone and prediction heads frozen.} (a) Rendering quality on RE10K scenes from identical context views. WorldMirror 2.0 (WM2.0) exhibits severe rendering artifacts, while RDGSplat achieves superior novel view synthesis. (b) The gain holds for three feed-forward models of two architectures, under both random and pretrained initialization. (c) RDGSplat optimizes 205.5\,M parameters against a frozen 1.4\,B backbone under photometric supervision alone.}
  \label{fig:teaser}
  \end{center}
}]

\maketitle
\begin{abstract}
Recent 3D foundation models extend to novel view synthesis by carrying a Gaussian head on the representation they already use for reconstruction. However, their rendering quality lags far behind their reconstruction accuracy, because the renderer reads a geometry and a representation that are both optimized for metric reconstruction. Prior methods alleviate this by updating the backbone weights, but they thereby sacrifice the metric predictions the model was built for. To this end, we propose RDGSplat, which decodes a geometry dedicated to rendering from a frozen 3D foundation model while keeping its metric predictions exact. In particular, we devise Render-Dedicated Geometry Decoding, which duplicates the pretrained decoders and optimizes the duplicates under photometric supervision alone. Then, a Target-Pose Conditioned Adapter is introduced to adapt the representation those decoders read towards rendering, conditioned on the predicted target pose. Extensive experiments show that RDGSplat improves novel view synthesis for three feed-forward models of two architectures on four benchmarks, with every pretrained weight frozen. On RE10K, it raises WM2.0 from 20.918 to 25.149\,dB by training 205.5\,M added parameters against a frozen 1.4\,B backbone, and leaves its predicted depth and pose unchanged.
\end{abstract}    
\section{Introduction}
\label{sec:intro}
Recovering the geometry of a scene from images is a long-standing goal in computer vision~\cite{schonberger2016structure, wang2025vggt, wang2024dust3r}. Classical Structure-from-Motion~\cite{schonberger2016structure, wang2024vggsfm} and Multi-View Stereo~\cite{schonberger2016pixelwise} attain high precision, but pay an iterative optimization cost for every new scene. Recent 3D foundation models such as DUSt3R~\cite{wang2024dust3r} and VGGT~\cite{wang2025vggt} instead amortize reconstruction across scenes~\cite{tang2025mvdust3r, yang2025fast3r, wang2024spann3r, wang2025continuous}, predicting depth, camera pose and pointmaps in a single feed-forward pass. Models such as HY-World 2.0~\cite{team2026hy} further carry a Gaussian head on the same representation, so that one network can serve both reconstruction and rendering in place of two systems.

Feed-forward novel view synthesis has meanwhile advanced along a separate line, regressing renderable primitives from sparse views under an image-quality objective. pixelSplat~\cite{charatan2024pixelsplat} and MVSplat~\cite{chen2024mvsplat} regress primitives from posed images, while NoPoSplat~\cite{ye2025no} needs no pose at inference and NAS3R~\cite{huang2026none} dispenses with pose supervision altogether. Optimized for image quality alone, these methods recover a scene only up to an unknown scale, and their objective leaves metric depth unconstrained, unlike dedicated depth estimators~\cite{piccinelli2024unidepth, yang2024depthanything}.

\begin{figure}[t]
  \centering
  \includegraphics[width=\linewidth]{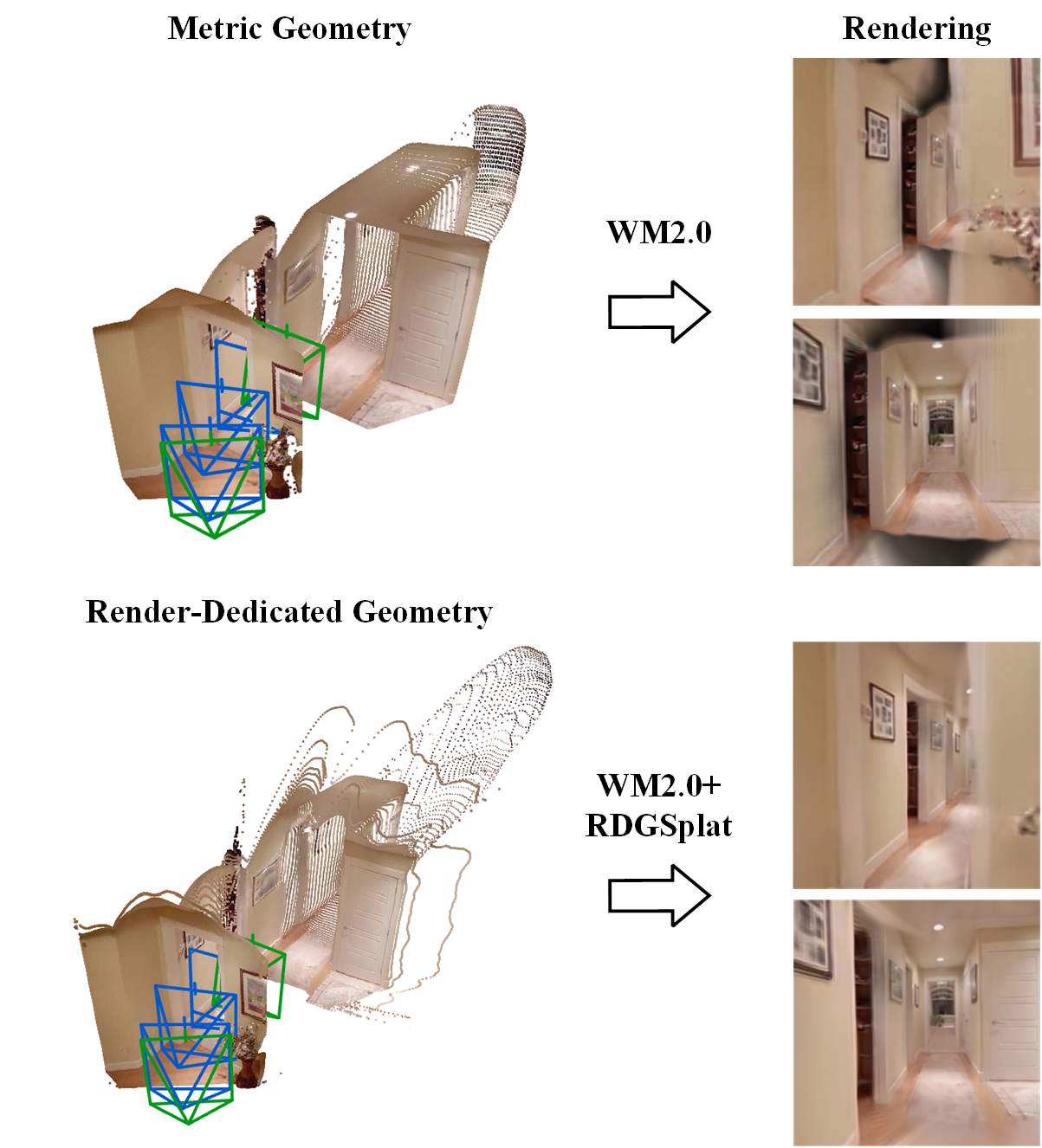}
  \caption{\textbf{The two geometries of a single scene.} Both are decoded from the same frozen backbone and the same context views. The metric geometry forms dense, solid surfaces, and the views rendered from it leave the image borders empty and blur the scene. The render-dedicated geometry spreads these surfaces into thin layers, from which the full corridor is rendered.}
  \label{fig:geometry}
  % \vspace{-0.3cm}
\end{figure}

However, a 3D foundation model renders far less accurately than it reconstructs. As Fig.~\ref{fig:teaser}(a) shows, frozen WM2.0 leaves large regions of the target view empty, which the same model carrying RDGSplat recovers. On RE10K, WM2.0 has the lowest PSNR in Table~\ref{tab:nvs}, yet surpasses DUSt3R and VGGT in depth and pose in Table~\ref{tab:metric}. Fig.~\ref{fig:geometry} traces this deficit to the geometry supplied to the renderer rather than to the renderer itself. This geometry is estimated under a metric objective rather than optimized for rendering. Such an objective rewards a surface placed where the scene is, leaving unconstrained how thick it is and where a primitive must sit to be seen from a new view. Moreover, the representation behind this geometry is computed for reconstruction without knowledge of the target view, so every target view is decoded from the same tokens.

Adaptations of foundation models for rendering~\cite{ye2025no, jiang2025anysplat} alleviate this by updating the backbone weights. Yet the metric heads and the renderer share these weights. Updating them therefore alters the metric predictions, and can degrade them~\cite{zhang2025emergent}. A practitioner needing measurement alongside synthesis must then run a dedicated renderer beside the foundation model, reinstating the two separate systems that a joint model was meant to replace.

The conflict arises because a geometry may be scored under either of two criteria. A \emph{metric geometry} is scored against the physical scene, whereas a \emph{render-dedicated geometry} is scored on novel view synthesis quality. The two criteria diverge~\cite{kerbl20233d, huang20242dgs, lu2024scaffoldgs}, since geometric accuracy does not ensure that primitives render well from a new view. Per-scene splatting reports the same tension, where geometric regularization costs image quality~\cite{ren20252dgs, huang2025gaussian}, a shared primitive attribute is split between the two objectives~\cite{li2025car, zhou2026geometry}, and photometric supervision is an ambiguous geometric signal~\cite{li2026revisiting}. A 3D foundation model holds only the metric geometry, and rendering needs the other as well.

% To this end, we propose RDGSplat, which decodes the render-dedicated geometry from a frozen model while the metric geometry is preserved intact, as illustrated in Fig.~\ref{fig:framework}. Instead of adapting one representation towards a compromise, RDGSplat evaluates the frozen backbone twice and assigns one evaluation to each criterion. The first is unmodified and continues to feed the pretrained heads, so the metric predictions are identical to the pretrained model's by construction. The second is adapted within the residual stream and conditioned on the target camera pose, and duplicated decoders read its tokens to estimate a render-dedicated geometry under photometric supervision alone. One model thereby serves both tasks at the cost of one extra evaluation.

% Because the render branch duplicates decoders the backbone already provides and the adapter operates inside its residual stream, the design depends on the interface of a backbone rather than on its weights. The same two modules integrate with three existing transformer-based 3D models without modification, and all three gain, spanning both random and pretrained initialization, as Fig.~\ref{fig:teaser}(b) shows.

To this end, we propose RDGSplat, which decodes the render-dedicated geometry from a frozen model while the metric geometry is preserved intact, as illustrated in Fig.~\ref{fig:framework}. Instead of updating the shared weights towards a compromise, RDGSplat leaves the pretrained heads and their input untouched, so that the metric predictions are identical to the pretrained model's by construction. Duplicated decoders estimate the render-dedicated geometry under photometric supervision alone. The representation they read is computed only from the context views and adapted towards rendering under the predicted target pose. One model thereby serves both tasks. Since the render branch duplicates existing decoders and the adapter operates inside the residual stream, the same modules integrate unmodified with three transformer-based 3D models of two architectures. As Fig.~\ref{fig:teaser}(b) shows, all three gain, under both random and pretrained initialization.

Our main contributions are as follows:

\begin{itemize}
\item We propose RDGSplat, a novel framework that equips a frozen 3D foundation model with a render-dedicated geometry, improving the views it renders while preserving its metric predictions exactly.
\item We devise Render-Dedicated Geometry Decoding (RDGD), which estimates this geometry through duplicated pretrained decoders optimized under photometric supervision alone, so that rendering quality, rather than a metric estimate, constrains the primitives.
\item A Target-Pose Conditioned Adapter (TCA) is introduced to adapt the frozen representation from reconstruction towards rendering, conditioned on the predicted target pose.
\item Experiments on four benchmarks show that RDGSplat, with every pretrained weight frozen, improves novel view synthesis for three feed-forward models of two architectures under random and pretrained initialization.
\end{itemize}
\section{Related Work}
\label{sec:related}

\subsection{3D Foundation Models}

Structure-from-Motion~\cite{schonberger2016structure, wang2024vggsfm, fu2024colmap} and Multi-View Stereo~\cite{schonberger2016pixelwise} estimate cameras and depth by iterative optimization, one scene at a time.

DUSt3R~\cite{wang2024dust3r} replaces that procedure for an image pair with one forward pass, regressing a pointmap in a common frame. MASt3R~\cite{leroy2024grounding} adds dense matching and MonST3R~\cite{zhang2024monst3r} admits dynamic content, while MV-DUSt3R+~\cite{tang2025mvdust3r} and Fast3R~\cite{yang2025fast3r} extend the pass to many views. VGGT~\cite{wang2025vggt} predicts depth, camera parameters and pointmaps together, and later models scale it~\cite{wang2026vggtomega}, remove its dependence on a reference frame~\cite{wang2026pi3}, or recover the visual space from arbitrary view counts~\cite{lin2025depth}. Other models add memory so that views may be processed as a stream, held as a growing token bank~\cite{wang2024spann3r, wu2025point3r}, a causal key-value cache~\cite{zhuo2025streaming}, or a fixed-size recurrent state~\cite{wang2025continuous}. Related estimators report depth in absolute units~\cite{piccinelli2024unidepth, yang2024depthanything}, from a single image or from many~\cite{keetha2026mapanything, wang2025amb3r}.

Throughout, the geometry is scored against annotated measurements. Some backbones are extended with a Gaussian head on the representation they already compute, among them HY-World 2.0~\cite{team2026hy}, so that one network may serve reconstruction and rendering together. The geometry that head reads is nonetheless still what the metric objective produced. Our work takes such a model as given and supplies the renderer with a geometry of its own.

\subsection{Feed-Forward Novel View Synthesis}

Early methods conditioned a radiance field on image features~\cite{yu2021pixelnerf} and inherited the slow volumetric rendering of NeRF~\cite{mildenhall2021nerf}. Gaussian splatting~\cite{kerbl20233d} replaced it with rasterized primitives, whose placement and extent later variants refine~\cite{huang20242dgs, lu2024scaffoldgs}. Building on it, pixelSplat~\cite{charatan2024pixelsplat} and MVSplat~\cite{chen2024mvsplat} regress per-pixel Gaussians from posed image pairs.

NoPoSplat~\cite{ye2025no} removes the pose requirement at inference, while PF3plat~\cite{hong2024pf3plat} and SelfSplat~\cite{kang2025selfsplat} predict pose and primitives in separate branches. SPFSplat~\cite{huang2025no, huang2025spfsplatv2} unifies both in one backbone, and NAS3R~\cite{huang2026none} dispenses with pose, depth and intrinsics supervision. Others discard explicit geometry for a latent representation~\cite{jiang2025rayzer, jin2024lvsm}, build on a frozen geometry predictor~\cite{zhang2026anchorsplat}, or decouple appearance from geometry~\cite{ren2026aesplat}. Pose-refining fields~\cite{lin2021barf} optimize cameras per scene, at the per-scene cost that feed-forward models are designed to avoid.

Because rendering is the only quantity scored, the geometry these methods hold is shaped to serve it, and most recover a scene only up to scale. Those reaching absolute units inherit them from a metric initialization~\cite{ye2025no, hong2024pf3plat} or distil a metric teacher~\cite{jiang2025anysplat}, and the scale therefore comes from a prior model rather than from the rendering objective. Measurement alongside synthesis thus still requires a second model, although the tasks that consume structure~\cite{yan2024gsslam, mao2025poma} depend on absolute units.

% \begin{figure*}[t]
%   \centering
%   \includegraphics[width=\linewidth]{imgs/FIG2-RDGSplat.png}
%   \caption{\textbf{Overall framework.} Given context images and a target camera pose, RDGSplat evaluates a frozen 3D foundation model twice and produces a metric geometry and a render-dedicated geometry. The unmodified evaluation feeds the pretrained heads, which predict the depth, camera poses and pointmaps that form the metric geometry. TCA takes Plucker rays, high-frequency content and the projected state as its condition, and adapts the second evaluation within the residual stream. RDGD decodes a render depth and per-primitive attributes from the adapted representation, which the rasterizer renders into the target view.}
%   \label{fig:framework}
% \end{figure*}

\begin{figure*}[t]
  \centering
  \includegraphics[width=\linewidth]{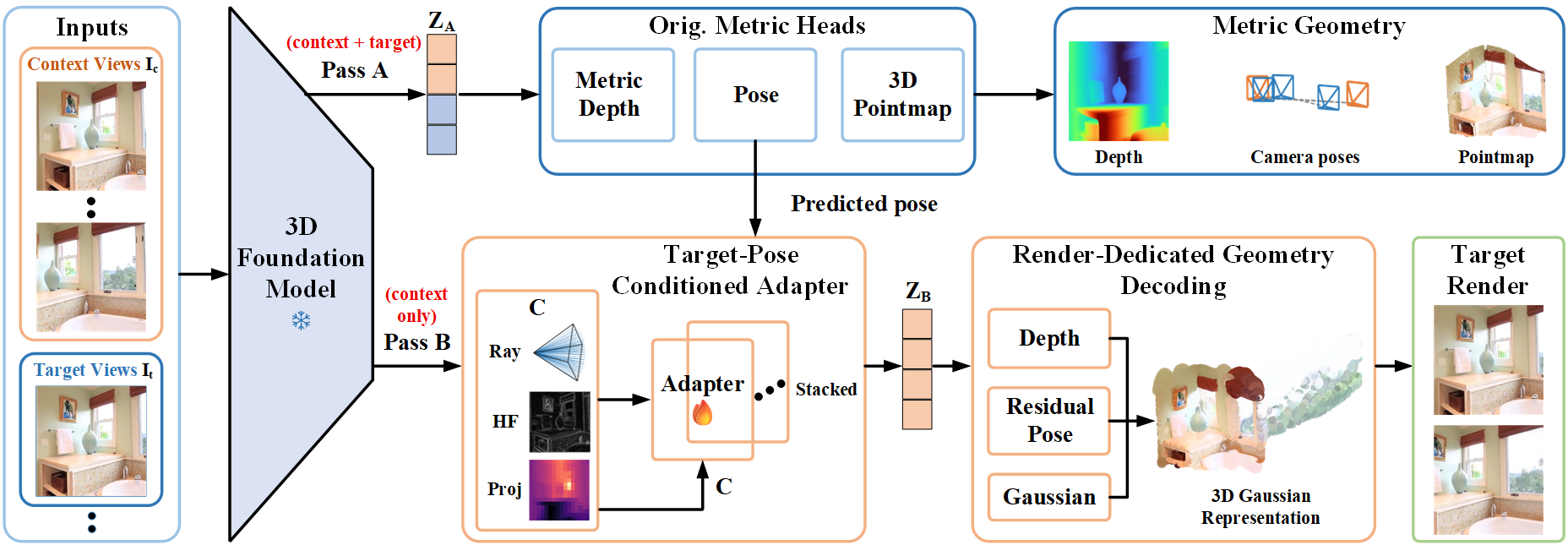}
  \caption{\textbf{Overall framework.} RDGSplat evaluates a frozen 3D foundation model twice and produces a metric geometry and a render-dedicated geometry. The token bars mark which views each evaluation carries: $z_A$ spans the context and target views, $z_B$ the context views alone. The unmodified evaluation feeds the pretrained heads, which predict the depth, camera poses and pointmaps forming the metric geometry. TCA takes Pl\"{u}cker rays, high-frequency content and the projected state as its condition, and adapts the second evaluation within the residual stream. RDGD decodes from $z_B$ a render depth and per-primitive attributes, which the rasterizer renders into the target view.}
  \label{fig:framework}
\end{figure*}

\subsection{Adapting Pretrained Backbones}

Self-supervised pretraining supplies the representation~\cite{dosovitskiy2021vit, oquab2023dinov2}, and in 3D vision cross-view completion~\cite{weinzaepfel2022croco, weinzaepfel2023crocov2} pretrains for geometry specifically. Adapter layers~\cite{houlsby2019parameter}, low-rank updates~\cite{hu2022lora} and prompt tuning~\cite{jia2022visual} each optimize a small set of added parameters against a frozen backbone. ControlNet~\cite{zhang2023adding} conditions a frozen diffusion model through a trainable copy of its encoding blocks, whose outputs rejoin the frozen path. Each of these keeps the pretrained weights intact.

In 3D vision, a frozen backbone may be extended with a decoder~\cite{liu2025monosplat, szymanowicz2025flash3d, zhao2026vg2gt}, a geometric adapter~\cite{peng2025omnivggt}, or a test-time update that trains no parameters~\cite{chen2026ttt3r, elflein2026vggt, xing2026adaptsplat}. Rendering, however, is most often reached by updating the weights instead, as in NoPoSplat~\cite{ye2025no} from MASt3R, AnySplat~\cite{jiang2025anysplat} from VGGT, and VGGT~\cite{wang2025vggt} itself. Updating resolves the conflict by abandoning one objective, and its cost is measurable: a full update degrades the dense reconstruction~\cite{zhang2025emergent}. Splatt3R~\cite{smart2024splatt3r} instead freezes MASt3R and offsets each predicted point, leaving the representation unadapted. We keep both objectives by giving each its own geometry from the same frozen weights, which asks more of an adapter than parameter efficiency: the pretrained output must remain exactly unchanged. None of the methods above is held to that requirement, because in each of them the adapted representation is the only one produced.

\section{Method}
\label{sec:method}

Existing 3D foundation models render from the geometry predicted by their metric heads, which limits the rendering quality in two aspects. First, this geometry is optimized for metric accuracy rather than for rendering. Second, the representation from which this geometry is decoded is computed for reconstruction, without knowledge of the target view. To address these issues, RDGSplat introduces RDGD for the former and TCA for the latter, as illustrated in Fig.~\ref{fig:framework}. Both operate on a second evaluation of the frozen backbone, while the first evaluation keeps the depth, pose and pointmap predictions of the pretrained model unchanged.

\subsection{Preliminary}
\label{sec:prelim}

\noindent\textbf{Task definition.} Let $\mathcal{I}_c = \{I_1,\ldots,I_N\}$ denote $N$ sparse context images of a static scene, where $N$ is small and typically two, and $I_t$ a target view of the same scene, rendered by the model as $\hat{I}_t$. No camera parameters are given. Whereas many methods take the target camera as an input that defines the query view~\cite{charatan2024pixelsplat, chen2024mvsplat, ye2025no}, we follow the pose-free practice~\cite{huang2026none, huang2025no} and feed the target view into the backbone to estimate its pose. Its pixels serve pose estimation (Sec.~\ref{sec:rdgd}), and are not decoded into primitives.

\noindent\textbf{Frozen backbone and its heads.} A multi-task 3D foundation model $f_\theta$ encodes the images into a token representation, from which pretrained heads $h_m$ decode metric geometry and a Gaussian head decodes primitives for rendering:
\begin{equation}
z_A = f_\theta(\mathcal{I}_c \cup \{I_t\}), \qquad (D, \Pi, X) = h_m(z_A)
\label{eq:backbone}
\end{equation}
where $z_A$ holds the tokens of $K$ patches per view, and $D$, $\Pi$ and $X$ are the predicted depth, camera parameters and pointmaps. The Gaussian head reads $z_c = f_\theta(\mathcal{I}_c)$, computed from the context images alone. The parameters $\theta$ are frozen throughout.

\noindent\textbf{Dual evaluation.} A naive approach is to fine-tune $\theta$ for rendering. However, $h_m$ reads tokens computed with the same $\theta$, and such fine-tuning will alter the metric predictions. We therefore keep $\theta$ frozen and introduce an adapted second evaluation:
\begin{equation}
z_B = f_\theta(\mathcal{I}_c \,;\, \mathrm{TCA})
\label{eq:dual}
\end{equation}
where $z_B$ is adapted over the context images $\mathcal{I}_c$ alone and replaces $z_c$ at the decoders, while $z_A$ remains unmodified. The predictions of $h_m$ are therefore identical to the pretrained model's. Since $z_B$ depends on the predicted target pose (Sec.~\ref{sec:tca}), the second evaluation is repeated per target view, so that the render-dedicated geometry is tailored to each target view.

\subsection{Render-Dedicated Geometry Decoding}
\label{sec:rdgd}

\begin{figure}[t]
  \centering
  \includegraphics[width=\linewidth]{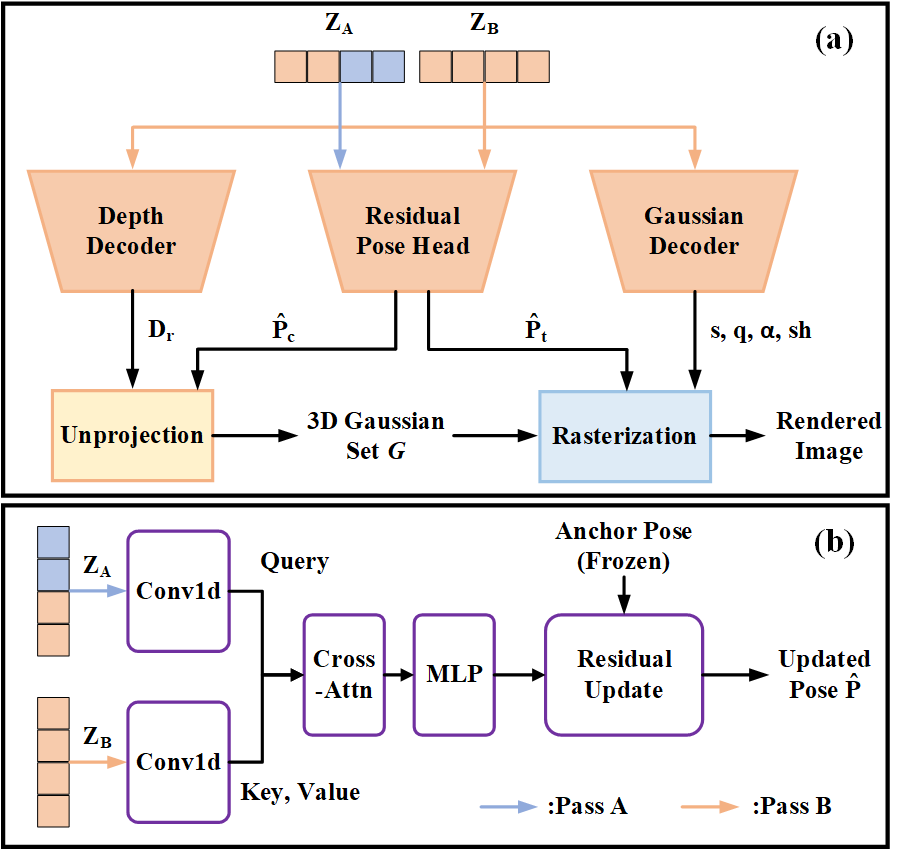}
  \caption{\textbf{Render-Dedicated Geometry Decoding.} (a) Duplicated depth and Gaussian decoders read $z_B$ and predict a render depth and per-primitive attributes, while the residual pose head reads both streams and corrects the poses; the primitives are then unprojected and rasterized. (b) The residual pose head projects both streams, attends from the joint stream to the adapted context stream, and updates a frozen anchor pose.}
  \label{fig:rdgd}
\end{figure}

Trained under metric supervision, the pretrained depth head predicts where a surface lies, but not how the primitives should be arranged to render it from a new view. As Fig.~\ref{fig:geometry} shows, the resulting geometry forms dense, solid surfaces, and the views rendered from it leave the image borders empty. Fine-tuning the pretrained decoders with a photometric loss is a direct remedy. Nevertheless, the depth head also produces the metric depth, which would then be sacrificed for rendering quality. Instead, we propose RDGD, which estimates a render-dedicated geometry with separate decoders and leaves the original ones unchanged.

\noindent\textbf{Duplicated decoders.} Specifically, RDGD duplicates two pretrained decoders, namely the depth head $g_d$ and the Gaussian decoder $g_s$, where the latter comprises the Gaussian feature head and its parameter decoder. Each duplicate is initialized with the parameters of its source and then optimized under photometric supervision, while the source remains frozen, so that no pretrained weight is updated. As shown in Fig.~\ref{fig:rdgd}(a), both duplicates take $z_B$ as input and predict a render depth and the per-primitive attributes:
\begin{equation}
D_r = g_d(z_B), \qquad (s, q, \alpha, sh) = g_s(z_B)
\label{eq:decoders}
\end{equation}
where $s$, $q$, $\alpha$ and $sh$ denote scale, rotation, opacity and view-dependent color. The render-dedicated geometry thus comprises not only where each primitive lies, but also its scale, rotation and opacity. Supervised by rendering quality alone, the duplicates are free to depart from the geometry they inherit, whereas the frozen depth head still provides the metric surface. The sources serve as initialization because both geometries describe the same observed surfaces.

\noindent\textbf{Residual pose head.} Rendering also depends on the camera poses, yet fine-tuning the camera head under photometric supervision will lead to severe degradation of target-view pose accuracy. This is because a primitive and a camera moving together along the viewing direction project to nearly the same pixels. We thus keep the pretrained pose as a fixed anchor and learn only a penalized residual on its translation. In Fig.~\ref{fig:rdgd}(b), the head projects the camera tokens of both streams and attends from $z_A$ to $z_B$ to regress a per-frame translation correction:
\begin{equation}
\begin{aligned}
q &= W_q z_A, \quad k = W_k z_B, \quad v = W_v z_B, \\
\Delta\tau &= \mathrm{MLP}\big( q + \mathrm{Attn}(q, k, v) \big), \\
\hat{P} &= (R_A, \; \tau_A + \Delta\tau)
\end{aligned}
\label{eq:pose}
\end{equation}
where $W_q$, $W_k$ and $W_v$ are per-frame projections applied to the camera token of each view, and $P_A = (R_A, \tau_A)$ is the pretrained pose. The rotation $R_A$ is kept unchanged. The queries cover both the context and target views, whereas the keys and values cover the context views only. The first context view defines the reference frame and receives no correction, so $\Delta\tau$ and Eq.~\ref{eq:total} cover the remaining views only.

\noindent\textbf{Assembly and rendering.} In the lower row of Fig.~\ref{fig:rdgd}(a), the render depth is unprojected into primitive centers with the corrected context poses $\hat{P}_c$. Together with the decoded attributes, these centers form the Gaussian set $G$, which is rasterized at the corrected target pose $\hat{P}_t$:
\begin{equation}
\begin{aligned}
\mu &= \pi^{-1}(D_r, \hat{P}_c), \quad G = \{\mu, s, q, \alpha, sh\}, \\
\hat{I}_t &= \mathrm{Rast}(G, \hat{P}_t)
\end{aligned}
\label{eq:assembly}
\end{equation}
where $\pi^{-1}$ denotes unprojection and $\mathrm{Rast}$ the voxel-merging rasterizer of the backbone.

%%%% ---- 3.3 Target-Pose Conditioned Adapter ----

\subsection{Target-Pose Conditioned Adapter}
\label{sec:tca}

\begin{figure}[t]
  \centering
  \includegraphics[width=\linewidth]{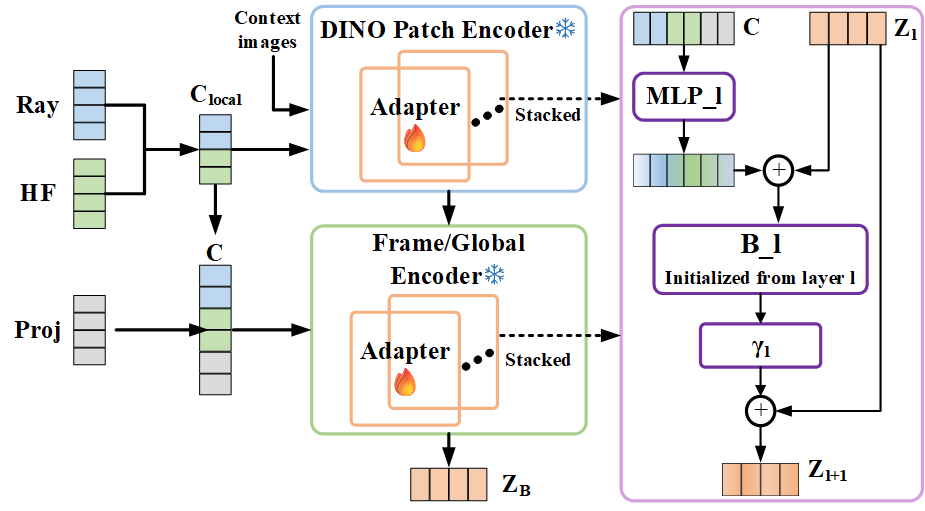}
  \caption{\textbf{Target-Pose Conditioned Adapter.} Pl\"{u}cker rays, high-frequency content and the projected state form the condition $C$, of which the per-frame patch encoder receives only the per-view prefix $C_{\text{local}}$. Each block $B_l$ is initialized from the layer it adapts, and enters the residual through a learned scale.}
  \label{fig:tca}
\end{figure}

RDGD re-estimates the primitives with new decoders, but the tokens these decoders read are still produced by the frozen backbone for reconstruction. Since $\theta$ also serves $h_m$, these tokens cannot be adapted by updating $\theta$. To this end, we propose TCA to adapt the representation within the second evaluation alone, as shown in Fig.~\ref{fig:tca}. As rendering serves a particular view, the adaptation is conditioned on the predicted target pose, and $z_B$ excludes the target image.

\noindent\textbf{Condition construction.} The condition $C$ consists of three terms. The Pl\"{u}cker ray coordinates $C_{\text{ray}}$ encode the viewing geometry of each context camera. The high-frequency content $C_{\text{hf}}$, i.e., the magnitude of the residual between each context image and its low-pass filtered version, indicates how much spatial detail the patch embedding attenuates. The projected state $C_{\text{prj}}$ is the only target-dependent term. It records where the primitive of each token appears in the target view, and is computed from the predicted poses together with a preliminary render depth $g_d(z_c)$. Its channels are the projected coordinates, in-frame and z-buffer visibility, two depth ratios, the cosine between both viewing rays and the primitive count per target pixel. $C_{\text{ray}}$ and $C_{\text{hf}}$, by contrast, describe each token within its own view. Each term is averaged per patch. The three terms are concatenated and lifted by a Fourier encoding $\psi$:
\begin{equation}
C = \psi\big( [\, C_{\text{ray}}, \; C_{\text{hf}}, \; C_{\text{prj}} \,] \big)
\label{eq:condition}
\end{equation}

\noindent\textbf{Adapter blocks and injection sites.} An adapter appended after the final backbone layer would pass its output straight to the decoders, with no pretrained layer left to transform it. TCA is instead injected into the residual stream at a set $S$ of depths, uniformly distributed over the per-frame patch encoder and the cross-view blocks, so that every subsequent layer transforms the adaptation. At each site $l \in S$, the condition is projected to the token width and added to the output of the frozen layer $L_l$. A block $B_l$ transforms this sum, and its output joins the residual stream through a learned per-channel scale:
\begin{equation}
\tilde{z}_l = L_l(z_l), \quad z_{l+1} = \tilde{z}_l + \gamma_l \odot B_l\big( \tilde{z}_l + \mathrm{MLP}_l(C) \big)
\label{eq:inject}
\end{equation}
where $L_l$ denotes the frozen layer $l$, the block $B_l$ is initialized from the pretrained layer $l$, internal residual connections included, and thereafter trained, and $\mathrm{MLP}_l$ maps the condition to the token width $d$. The scale $\gamma_l \in \mathbb{R}^d$ is initialized to $10^{-3}$, so that the adapter branch starts close to the identity. Adding the condition rather than concatenating it keeps $B_l$ at its original width and thus able to inherit the pretrained parameters of layer $l$.

As the patch encoder processes each view independently, $C_{\text{prj}}$ is undefined at that depth. The sites within the patch encoder receive only $C_{\text{local}} = \psi([C_{\text{ray}}, C_{\text{hf}}])$, while the sites within the cross-view blocks receive the full condition.

%%%% ---- 3.4 Training Objectives ----

\subsection{Training Objectives}
\label{sec:objectives}

Both modules are optimized jointly against held-out views, with every parameter of $f_\theta$ and $h_m$ frozen. No annotation exists for a geometry defined by how it renders, so the photometric objective alone drives the duplicated decoders.

\noindent\textbf{Photometric objective.} The rendered view $\hat{I}_t$ is supervised against $I_t$ by a pixel and a perceptual term:
\begin{equation}
\mathcal{L}_{\text{photo}} =
\mathcal{L}_{\text{mse}}(\hat{I}_t, I_t) +
\lambda_p \cdot \mathcal{L}_{\text{lpips}}(\hat{I}_t, I_t)
\label{eq:photo}
\end{equation}
where $\mathcal{L}_{\text{mse}}$ is the mean squared error over pixels and $\mathcal{L}_{\text{lpips}}$ a perceptual distance, weighted by $\lambda_p$ = 0.05. As no term requires geometric annotation, the same objective applies to every backbone, regardless of how it was pretrained.

\noindent\textbf{Pose regularization.} The correction $\Delta\tau$ of Sec.~\ref{sec:rdgd} is penalized relative to the anchor translation, which lets one weight suit every backbone:
\begin{equation}
\mathcal{L} = \mathcal{L}_{\text{photo}} +
\lambda_r \cdot \left\| \frac{\Delta\tau}{\|\tau_A\|} \right\|^2
\label{eq:total}
\end{equation}
where $\lambda_r$ = 1.

\section{Experiments}
\label{sec:exp}

%%%% ---- 4.1 Experimental Setup ----

\subsection{Experimental Setup}
\label{sec:setup}

\noindent\textbf{Datasets and baselines.} We train on RE10K~\cite{zhou2018stereo} and test zero-shot on ACID~\cite{liu2021infinite}, DL3DV~\cite{ling2024dl3dv} and the object-centric DTU~\cite{jensen2014large}. Metric geometry is evaluated on KITTI~\cite{geiger2012we} for video depth and on RE10K for two-view pose. Because RDGSplat attaches to a pretrained model rather than replacing one, we report two settings for each backbone: the pretrained model evaluated on its own, and the same model with RDGSplat attached. For synthesis we further compare against pixelSplat~\cite{charatan2024pixelsplat}, MVSplat~\cite{chen2024mvsplat}, NoPoSplat~\cite{ye2025no} and NAS3R~\cite{huang2026none}, while for metric geometry we compare against DUSt3R~\cite{wang2024dust3r}, MASt3R~\cite{leroy2024grounding} and VGGT~\cite{wang2025vggt}. Following NAS3R~\cite{huang2026none}, we adopt its evaluation indices, view sampling and image resolution throughout.

\noindent\textbf{Implementation details.} RDGSplat is implemented in PyTorch and trained on a single NVIDIA A6000. Every run reported in this paper, including the ablation of Sec.~\ref{sec:ablation}, draws on one random 20\% sample of the RE10K training split. The two modules are optimized jointly under the objective of Sec.~\ref{sec:objectives}, for 200k iterations on WM2.0 and 50k on both NAS3R variants. Synthesis is measured at three levels by pixel-level PSNR, patch-level SSIM~\cite{wang2004image} and feature-level LPIPS~\cite{zhang2018unreasonable}. Depth is measured by the absolute relative error and the inlier ratio at 1.25. Pose is measured by the area under the cumulative error curve at 10 and 20 degrees. Target views are rendered from predicted rather than ground-truth poses, and the input is two context views unless stated otherwise.

\subsection{Comparison with State-of-the-Art Methods}
\label{sec:sota}

\begin{table*}[htbp]
\caption{\textbf{Two-view novel view synthesis.} Self-sup. marks training without geometric annotation from the initialization in Init., which for RDGSplat rows applies to the added modules. The best results among self-supervised methods under each initialization are highlighted. $\dagger$ indicates variants reproduced from officially released codebase and weights.}
\label{tab:nvs}
\centering
\setlength{\tabcolsep}{6pt}
\renewcommand{\arraystretch}{1.2}
\resizebox{\linewidth}{!}{
\begin{tabular}{lcc|ccc|ccc|ccc|ccc}
\toprule[1pt]
\multirow{2}{*}{\textbf{Method}} & \multirow{2}{*}{\textbf{Init.}} & \multirow{2}{*}{\textbf{Self-sup.}}
& \multicolumn{3}{c|}{\textbf{RE10K}} & \multicolumn{3}{c|}{\textbf{ACID}}
& \multicolumn{3}{c|}{\textbf{DL3DV}} & \multicolumn{3}{c}{\textbf{DTU}} \\
\cmidrule(lr){4-6} \cmidrule(lr){7-9} \cmidrule(lr){10-12} \cmidrule(lr){13-15}
& & & PSNR $\uparrow$ & SSIM $\uparrow$ & LPIPS $\downarrow$
    & PSNR $\uparrow$ & SSIM $\uparrow$ & LPIPS $\downarrow$
    & PSNR $\uparrow$ & SSIM $\uparrow$ & LPIPS $\downarrow$
    & PSNR $\uparrow$ & SSIM $\uparrow$ & LPIPS $\downarrow$ \\
\midrule[0.6pt]
\midrule[0.6pt]
pixelSplat~\cite{charatan2024pixelsplat} & random & -- & 23.859 & 0.808 & 0.184 & 25.477 & 0.770 & 0.207 & 21.370 & 0.713 & 0.250 & 15.067 & 0.539 & 0.341 \\
MVSplat~\cite{chen2024mvsplat}    & random & -- & 24.012 & 0.812 & 0.175 & 25.525 & 0.773 & 0.199 & 20.344 & 0.673 & 0.258 & 14.542 & 0.537 & 0.324 \\
NoPoSplat~\cite{ye2025no}  & random & -- & 21.382 & 0.709 & 0.266 & 23.193 & 0.667 & 0.274 & 18.235 & 0.518 & 0.364 & 15.113 & 0.455 & 0.450 \\
NAS3R$^\dagger$~\cite{huang2026none} & random & \checkmark & 23.594 & 0.766 & 0.188 & 25.030 & 0.735 & 0.209 & 17.646 & 0.466 & 0.375 & 15.229 & 0.524 & 0.317 \\
\rowcolor{gray!12}
NAS3R + RDGSplat & random & \checkmark & \textbf{24.906} & \textbf{0.791} & \textbf{0.170} & \textbf{26.343} & \textbf{0.763} & \textbf{0.185} & \textbf{19.077} & \textbf{0.505} & \textbf{0.342} & \textbf{15.461} & \textbf{0.532} & \textbf{0.307} \\
\midrule[0.6pt]
PF3plat~\cite{hong2024pf3plat} & UniDepth v2 & \checkmark & 21.042 & 0.739 & 0.233 & 20.726 & 0.610 & 0.308 & -- & -- & -- & -- & -- & -- \\
NAS3R (VGGT)$^\dagger$~\cite{huang2026none} & VGGT & \checkmark & 26.394 & 0.867 & 0.130 & 26.878 & 0.812 & 0.163 & 19.830 & 0.628 & 0.274 & 16.961 & 0.632 & 0.242 \\
\rowcolor{gray!12}
NAS3R + RDGSplat & VGGT & \checkmark & \textbf{27.732} & \textbf{0.878} & \textbf{0.121} & \textbf{27.990} & \textbf{0.828} & \textbf{0.148} & \textbf{21.272} & \textbf{0.660} & \textbf{0.254} & \textbf{17.156} & \textbf{0.639} & \textbf{0.233} \\
\midrule[0.6pt]
WM2.0~\cite{team2026hy} & WM2.0 & -- & 20.918 & 0.747 & 0.210 & 21.752 & 0.679 & 0.266 & 16.674 & 0.576 & 0.319 & 14.445 & 0.460 & 0.338 \\
\rowcolor{gray!12}
WM2.0 + RDGSplat & WM2.0 & \checkmark & \textbf{25.149} & \textbf{0.831} & \textbf{0.149} & \textbf{25.559} & \textbf{0.771} & \textbf{0.183} & \textbf{20.361} & \textbf{0.647} & \textbf{0.272} & \textbf{17.104} & \textbf{0.567} & \textbf{0.284} \\
\bottomrule[1pt]
\end{tabular}
}
% \vspace{-0.3cm}
\end{table*}

\begin{table}[htbp]
\caption{\textbf{Metric geometry.} Video depth follows the CUT3R protocol with per-sequence scale alignment~\cite{wang2025continuous}, and two-view pose the AUC protocol of NAS3R~\cite{huang2026none}. The best and second-best results are highlighted.}
\label{tab:metric}
\centering
\setlength{\tabcolsep}{8pt}
\renewcommand{\arraystretch}{1.2}
%%%% Rows omitted: these methods report only one of the two metric groups.
%%%% depth only : MonST3R-GA 0.168/74.4, Spann3R 0.198/73.7, CUT3R 0.118/88.1,
%%%%              Point3R 0.136/84.2, StreamVGGT 0.173/72.1
%%%% pose  only : NoPoSplat 72.8/83.3, PF3plat 39.8/61.3, SPFSplat 75.5/84.5,
%%%%              SPFSplatV2 77.6/86.3, NAS3R (VGGT) 77.2/86.6
\resizebox{\linewidth}{!}{
\begin{tabular}{lcc|cc}
\toprule[1pt]
\multirow{2}{*}{\textbf{Method}} & \multicolumn{2}{c|}{\textbf{Video depth (KITTI)}} & \multicolumn{2}{c}{\textbf{Two-view pose (RE10K)}} \\
\cmidrule(lr){2-3} \cmidrule(lr){4-5}
 & Abs Rel $\downarrow$ & $\delta<1.25$ $\uparrow$ & $@10^\circ$ $\uparrow$ & $@20^\circ$ $\uparrow$ \\
\midrule[0.6pt]
DUSt3R~\cite{wang2024dust3r} & 0.144 & 81.3 & \underline{54.1} & \underline{70.2} \\
MASt3R~\cite{leroy2024grounding} & 0.183 & 74.5 & 49.4 & 67.1 \\
VGGT~\cite{wang2025vggt} & \underline{0.061} & \underline{97.0} & 47.4 & 65.8 \\
\rowcolor{gray!12}
WM2.0 (w/ and w/o RDGSplat) & \textbf{0.060} & \textbf{97.4} & \textbf{71.6} & \textbf{82.9} \\
\bottomrule[1pt]
\end{tabular}
}
% \vspace{-0.3cm}
\end{table}

Table~\ref{tab:nvs} presents the quantitative results across RE10K, ACID, DL3DV and DTU, under both random and pretrained initialization. It is shown that RDGSplat improves every backbone on all four benchmarks and under all three metrics, while updating no pretrained weight.

The largest gain appears on WM2.0, where the PSNR on RE10K rises by 4.231\,dB. We attribute this to its metric-supervised geometry, which departs furthest from what rendering requires and therefore benefits most from a render-dedicated geometry. With RDGSplat attached, WM2.0 also outperforms pixelSplat and MVSplat on RE10K and DTU, although both are trained for rendering end to end and receive ground-truth poses at inference. Besides, the gain persists where the backbone is already strong: on the VGGT initialization, the best baseline on RE10K before RDGSplat is attached, it still adds 1.338\,dB.

The gains also hold outside the training distribution. On ACID, DL3DV and DTU, all unseen during training, RDGSplat adds 3.807, 3.687 and 2.659\,dB to WM2.0, and 1.112, 1.442 and 0.195\,dB to the VGGT variant. The gain thus does not arise from fitting the training scenes.

\subsection{Preservation of Metric Geometry}
\label{sec:preserve}

Table~\ref{tab:metric} presents the metric geometry, produced by the frozen depth and camera heads of WM2.0 while RDGSplat is attached. WM2.0 shares one row with and without RDGSplat. It is shown that the joint model retains the metric accuracy of WM2.0, which leads on both axes with 0.060 Abs Rel and 71.6 AUC at $10^\circ$, against 0.061 for VGGT and 54.1 for DUSt3R.

Each predicted frame on the depth benchmark is identical to the pretrained model's, and the pose AUC agrees to four decimal places. One model thus serves both measurement and rendering, with no dedicated renderer beside it. Please refer to Supp.~Sec.~4 for the render-branch pose.

\begin{table}[tbp]
\caption{\textbf{Ablation studies of each component.} All rows are trained for 30k iterations, and Row 1 denotes WM2.0 with neither module attached. The best results are highlighted.}
\label{tab:ablation}
\centering
\setlength{\tabcolsep}{8pt}
\renewcommand{\arraystretch}{1.2}
%%%% DL3DV columns commented out
%%%% row 1: 16.674 0.576 0.319 | row 2: 18.483 0.600 0.305
%%%% row 3: 18.358 0.592 0.315 | row 4: 19.646 0.627 0.282
\resizebox{\linewidth}{!}{
\begin{tabular}{ccc|ccc|ccc}
\toprule[1pt]
& \multirow{2}{*}{\textbf{RDGD}} & \multirow{2}{*}{\textbf{TCA}}
& \multicolumn{3}{c|}{\textbf{RE10K}} & \multicolumn{3}{c}{\textbf{ACID}} \\
\cmidrule(lr){4-6} \cmidrule(lr){7-9}
& & & PSNR $\uparrow$ & SSIM $\uparrow$ & LPIPS $\downarrow$
    & PSNR $\uparrow$ & SSIM $\uparrow$ & LPIPS $\downarrow$ \\
\midrule[0.6pt]
1 & --          & --          & 20.918 & 0.747 & 0.210 & 21.752 & 0.679 & 0.266 \\
2 & \checkmark   & --          & 23.485 & 0.801 & 0.175 & 24.145 & 0.740 & 0.216 \\
3 & --          & \checkmark   & 23.032 & 0.788 & 0.186 & 23.992 & 0.736 & 0.218 \\
\rowcolor{gray!12}
4 & \checkmark   & \checkmark   & \textbf{24.499} & \textbf{0.819} & \textbf{0.160} & \textbf{24.941} & \textbf{0.760} & \textbf{0.197} \\
\bottomrule[1pt]
\end{tabular}
}
% \vspace{-0.3cm}
\end{table}

\subsection{Ablation Study on Component Design}
\label{sec:ablation}

% To assess the contribution of each module, we conduct ablation studies on RE10K and ACID using WM2.0 as the baseline. As shown in Table~\ref{tab:ablation}, when we only add RDGD to the baseline, it achieves 23.485\,dB on RE10K and 24.145\,dB on ACID. This indicates that decoding the primitives under photometric supervision, rather than taking them from the metric estimate, already improves rendering. Adding TCA alone, with the decoders kept at their pretrained weights, obtains 23.032 and 23.992\,dB, respectively. This shows that adapting the representation is effective on its own.

% When both modules are integrated into the framework, RDGSplat achieves the best performance with 24.499 and 24.941\,dB. Their separate gains over the baseline, 2.567 and 2.114\,dB on RE10K, sum to more than the 3.581\,dB they attain together, as do 2.393 and 2.240\,dB against 3.189\,dB on ACID. This shows that the two modules overlap in part, as both alter the primitives. Further, RDGD relies on inheriting the source parameters, with 23.485\,dB against 17.449\,dB on RE10K from random initialization (Supp.~Sec.~2), and most of the TCA gain comes from the condition rather than from the added capacity, which raises the blocks $B_l$ from 21.520 to 23.032\,dB (Supp.~Sec.~3).

To assess the contribution of each module, we conduct ablation studies on RE10K and ACID using WM2.0 as the baseline. As shown in Table~\ref{tab:ablation}, adding RDGD alone raises every metric on both datasets. This indicates that how the primitives are located, which the metric objective does not constrain, limits the rendering quality of the frozen model. Adding TCA alone, with the primitives still decoded by the pretrained decoders, also raises every metric. We attribute this to the adaptation of the tokens, which the frozen backbone otherwise computes identically for every target view.

When both modules are integrated into the framework, RDGSplat achieves the best performance, improving the baseline by 3.581 and 3.189\,dB. This demonstrates the effectiveness of addressing both the geometry supplied to the renderer and the representation it is decoded from. For RDGD, starting the duplicated decoders from random weights drops the quality below that of the frozen backbone, and fine-tuning only the pretrained Gaussian decoder leaves it well below RDGSplat. Both results show that the duplicates rely on the solution their sources already carry, and that a render-dedicated geometry needs its own depth and pose as well (Supp.~Sec.~2). For TCA, removing the condition from the adapter loses most of its gain. We assume that it is because the condition carries the predicted target pose, which the blocks alone cannot supply (Supp.~Sec.~3).

\subsection{Efficiency Analysis}
\label{sec:efficiency}

\begin{table}[tbp]
\caption{\textbf{Training efficiency.} A dash marks a quantity that does not apply to a frozen baseline.}
\label{tab:cost}
\centering
\setlength{\tabcolsep}{8pt}
\renewcommand{\arraystretch}{1.2}
%%%% Steps column removed: WM2.0+RDGSplat 200k, NAS3R+RDGSplat 50k
\resizebox{\linewidth}{!}{
\begin{tabular}{lcccc|c}
\toprule[1pt]
\textbf{Method} & \textbf{\shortstack{Pretrained\\weights updated}} & \textbf{RDGD} & \textbf{TCA} & \textbf{Trainable} & \textbf{Share} \\
\midrule[0.6pt]
WM2.0~\cite{team2026hy} & 0 & -- & -- & -- & -- \\
NAS3R~\cite{huang2026none} & all & -- & -- & 1.4\,B & 100\% \\
\rowcolor{gray!12}
WM2.0 + RDGSplat & \textbf{0} & 67.2\,M & 138.2\,M & 205.5\,M & 14.7\% \\
\rowcolor{gray!12}
NAS3R + RDGSplat & \textbf{0} & 67.1\,M & 138.2\,M & 205.3\,M & 14.7\% \\
\bottomrule[1pt]
\end{tabular}
}
\vspace{-0.1cm}
\end{table}

\begin{table}[tbp]
\caption{\textbf{Performance on novel view synthesis across different numbers of context views.} RDGSplat rows use one model trained across two to ten views, separate from that of Table~\ref{tab:nvs}. The best results are highlighted.}
\label{tab:views}
\centering
\setlength{\tabcolsep}{8pt}
\renewcommand{\arraystretch}{1.2}
\resizebox{\linewidth}{!}{
\begin{tabular}{ll|ccc}
\toprule[1pt]
\textbf{Views} & \textbf{Method} & PSNR $\uparrow$ & SSIM $\uparrow$ & LPIPS $\downarrow$ \\
\midrule[0.6pt]
\midrule[0.6pt]
\multirow{2}{*}{2} & WM2.0 & 20.918 & 0.747 & 0.210 \\
& WM2.0 + RDGSplat & \textbf{24.195} & \textbf{0.812} & \textbf{0.165} \\
\midrule[0.6pt]
\multirow{2}{*}{3} & WM2.0 & 22.592 & 0.792 & 0.179 \\
& WM2.0 + RDGSplat & \textbf{26.128} & \textbf{0.856} & \textbf{0.130} \\
\midrule[0.6pt]
\multirow{2}{*}{5} & WM2.0 & 23.796 & 0.822 & 0.157 \\
& WM2.0 + RDGSplat & \textbf{28.095} & \textbf{0.896} & \textbf{0.097} \\
\midrule[0.6pt]
\multirow{2}{*}{10} & WM2.0 & 24.540 & 0.846 & 0.141 \\
& WM2.0 + RDGSplat & \textbf{29.692} & \textbf{0.920} & \textbf{0.076} \\
\bottomrule[1pt]
\end{tabular}
}
\vspace{-0.1cm}
\end{table}

% Table~\ref{tab:cost} presents the training cost. It is shown that RDGSplat optimizes 205.5\,M parameters, 14.7\% of the 1.4\,B backbone, and updates none of its pretrained weights, whereas fine-tuning the same backbone for rendering requires all 1.4\,B. Besides, only 39.0\,M of those parameters are instantiated from scratch, 2.8\% of the backbone, while the remaining 166.5\,M duplicate modules it already carries. This shows that the render-dedicated geometry costs a duplicate of capacity already present rather than new capacity, whose decoders are therefore initialized from a solution to a related problem rather than from noise.

Table~\ref{tab:cost} presents the training cost. It is shown that RDGSplat optimizes 205.5\,M parameters, 14.7\% of the 1.4\,B backbone, and updates none of its pretrained weights, whereas fine-tuning the same backbone for rendering requires all 1.4\,B. Besides, only 39.1\,M of those parameters are instantiated from scratch, 2.8\% of the backbone, while the remaining 166.4\,M duplicate modules it already carries. This shows that the render-dedicated geometry builds on the geometry already pretrained in the backbone, rather than on new capacity. At inference, RDGSplat computes a preliminary render depth and repeats the second evaluation per target view. This raises the render latency from 262.3 to 688.5\,ms at two context views, while peak memory rises by at most 0.06\,GB. This overhead stays below one second per target view (Supp.~Sec.~5).

\subsection{Scaling to More Context Views}
\label{sec:views}

Table~\ref{tab:views} presents novel view synthesis on RE10K as the number of context views increases from two to ten, using one model trained across these counts (Supp.~Sec.~1). It is shown that RDGSplat surpasses the frozen backbone under every metric at every count, and that three context views already exceed the quality the backbone attains from ten under all three metrics, with 26.128 against 24.540\,dB. Besides, the margin widens steadily from 3.277\,dB at two views through 3.536 and 4.299\,dB at three and five to 5.152\,dB at ten. This suggests that the render-dedicated geometry converts each additional view into rendering quality more effectively than the metric geometry, so that the two sources of improvement compound rather than overlap.

\subsection{Qualitative Comparison}
\label{sec:qualitative}

\begin{figure}[t]
\centering
\includegraphics[width=\linewidth]{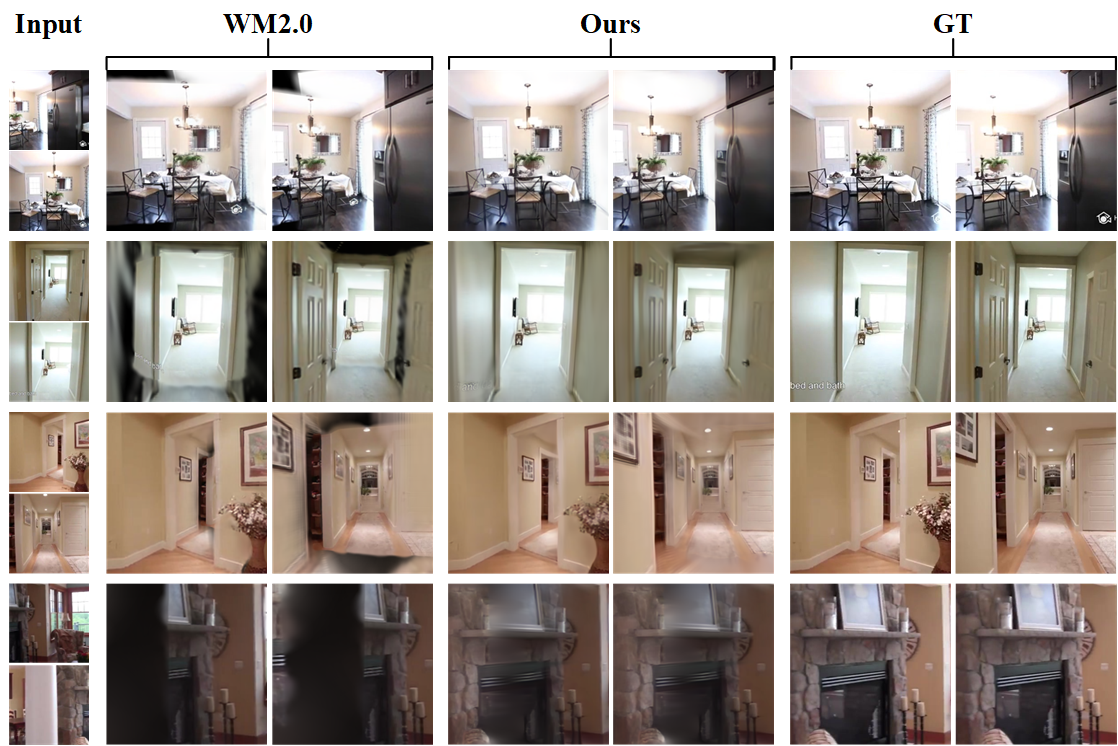}
\caption{\textbf{Qualitative comparison on RE10K.} Scenes sampled at random
from the test split. The leftmost column shows the two context views, and two
held-out target views follow.}
\label{fig:qualitative}
\end{figure}

Qualitative comparisons are provided in Fig.~\ref{fig:qualitative}. It is clear that WM2.0 leaves the image border empty and blurs the surfaces it does render. In the second scene, for instance, black streaks run along the frame edges, and in the fourth the left half of the fireplace becomes a dark smear. Instead, our method renders the full frame and closely follows the structure and color of the ground truth, including on the scenes where the frozen backbone degrades most. These comparisons show that RDGSplat renders more complete and more faithful target views than the frozen backbone, both at the image border and on the surfaces within the frame, in every scene shown. Please refer to Supp.~Sec.~6 for more qualitative results.

\section{Conclusion}
\label{sec:conclusion}
In this work, we present RDGSplat, a framework that improves the rendering of frozen 3D foundation models. Its key contribution is to separate a render-dedicated geometry from the metric one within a single frozen model. Duplicated decoders estimate this geometry under rendering supervision alone, and a target-pose conditioned adapter tailors their input to the target view. One model thus serves rendering and reconstruction at once, without geometric annotation and without updating any pretrained weight. Extensive experiments show that RDGSplat improves every backbone on four benchmarks while leaving its metric predictions unchanged, with the largest gains on WM2.0, the backbone with metric-supervised geometry.

\noindent\textbf{Limitations.} Our method assumes a transformer backbone with a pretrained depth head. Tuning camera translation directly under photometric supervision degrades pose accuracy, so the pretrained pose is still needed as an anchor.

% \newpage
% \input{sec/5b_llm_statement}

{
    \small
    \bibliographystyle{ieeenat_fullname}
    \bibliography{main}
}


\clearpage
\twocolumn[
  \begin{center}
    {\LARGE \textbf{Supplementary Material}}
    \vspace{1em}
  \end{center}
]

\setcounter{section}{0}
\setcounter{figure}{0}
\setcounter{table}{0}

%%%%%%%%%%%%%%%%%%%%%%%%%%%%%%%%%%%%%%%%%%%%%%%%%%%%%%%%%%%%%%%%%%%%%%%%%%%%%%
%%%% A. Implementation Details
%%%%%%%%%%%%%%%%%%%%%%%%%%%%%%%%%%%%%%%%%%%%%%%%%%%%%%%%%%%%%%%%%%%%%%%%%%%%%%

\section{Implementation Details}
\label{sec:impl}

Algorithm~\ref{alg:twopass} describes one inference of RDGSplat for a target view, in which the frozen backbone is evaluated twice. The first evaluation takes the context and target views and feeds the pretrained heads, whose poses, together with a preliminary render depth $g_d(z_c)$ on the context-only stream $z_c = f_\theta(\mathcal{I}_c)$ that the backbone itself computes, enter the condition. The second evaluation takes the context views alone with TCA active, and the duplicated decoders read its tokens. The condition is detached before it enters the second evaluation, so that $g_d$ is trained only through the render depth and not through the condition it helps to form. The first evaluation involves no TCA and no trainable parameter, so its predictions are those of the pretrained model. Table~\ref{tab:arch} below lists the quantities the main paper states in symbols.

The model of Table~5 of the main paper is trained separately from the two-view model of its Table~1, following the multi-view protocol of NAS3R~\cite{huang2026none}. Each training sample carries ten context views and one target view. The two outermost context views are placed by the view sampler of NAS3R, with a frame gap that widens from 25 to between 45 and 150 over the first 15k iterations. The other eight are drawn without repetition between them, and the target view lies within the same interval. At every iteration, the first and last context views are retained together with a random subset of the eight in between, whose size is uniform from zero to eight. One run therefore covers every count from two to ten context views, and the same weights are evaluated at each count of Table~5. Training starts from the pretrained WM2.0, as for the two-view model, and runs for 30k iterations with a batch size of two, under the objective of Sec.~3.4 of the main paper.

\begin{algorithm}[htbp]
\caption{One inference of RDGSplat for a target view.}
\label{alg:twopass}
\begin{algorithmic}[1]
\Require context images $\mathcal{I}_c$, target image $I_t$
\Ensure metric geometry $(D, \Pi, X)$, rendered target view $\hat{I}_t$
\State $z_A \gets f_\theta(\mathcal{I}_c \cup \{I_t\})$ \Comment{first evaluation (Pass A), unmodified}
\State $z_c \gets f_\theta(\mathcal{I}_c)$ \Comment{context-only stream of the same evaluation}
\State $(D, \Pi, X) \gets h_m(z_A)$ \Comment{pretrained heads}
\State $C \gets \psi([C_\text{ray}, C_\text{hf}, C_\text{prj}])$ \Comment{$C_\text{prj}$ from $g_d(z_c)$ and $\Pi$, detached}
\State $z_B \gets f_\theta(\mathcal{I}_c \,;\, \mathrm{TCA}(C))$ \Comment{second evaluation (Pass B), context views only}
\State $D_r \gets g_d(z_B)$
\State $(s, q, \alpha, sh) \gets g_s(z_B)$
\State $\Delta\tau \gets \mathrm{MLP}(W_q z_A + \mathrm{Attn}(W_q z_A, W_k z_B, W_v z_B))$
\State $\hat{P} \gets (R_A, \tau_A + \Delta\tau)$ \Comment{rotation carried over}
\State $\mu \gets \pi^{-1}(D_r, \hat{P}_c)$
\State $G \gets \{\mu, s, q, \alpha, sh\}$
\State $\hat{I}_t \gets \mathrm{Rast}(G, \hat{P}_t)$
\end{algorithmic}
\end{algorithm}

\begin{table}[htbp]
\caption{\textbf{Architecture detail.} The same four injection depths are used in the patch encoder and in the cross-view blocks. The four RDGD entries sum to the 67.2\,M of Table~4 of the main paper, up to rounding.}
\label{tab:arch}
\centering
\setlength{\tabcolsep}{8pt}
\renewcommand{\arraystretch}{1.2}
\resizebox{\linewidth}{!}{
\begin{tabular}{lc}
\toprule[1pt]
\textbf{Component} & \textbf{Value} \\
\midrule[0.6pt]
\midrule[0.6pt]
Injection sites $|S|$ & 4 in the patch encoder, 4 in the cross-view blocks \\
Site placement        & blocks 4, 11, 17, 23 \\
Token width $d$       & 1024 \\
Condition width       & 153 (cross-view blocks), 81 (patch encoder) \\
$\mathrm{MLP}_l$      & $153$ or $81 \rightarrow 4096 \rightarrow 1024$ \\
Fourier bands         & 4 \\
Block $B_l$         & copy of layer $l$ \\
Scale $\gamma_l$ init       & $10^{-3}$ \\
$g_d$                 & 32.65\,M \\
$g_s$                 & 32.67\,M feature, 0.30\,M parameter decoder \\
Residual pose head    & 1.62\,M \\
\bottomrule[1pt]
\end{tabular}
}
\end{table}

\begin{figure*}[htbp]
  \centering
  \includegraphics[width=\linewidth]{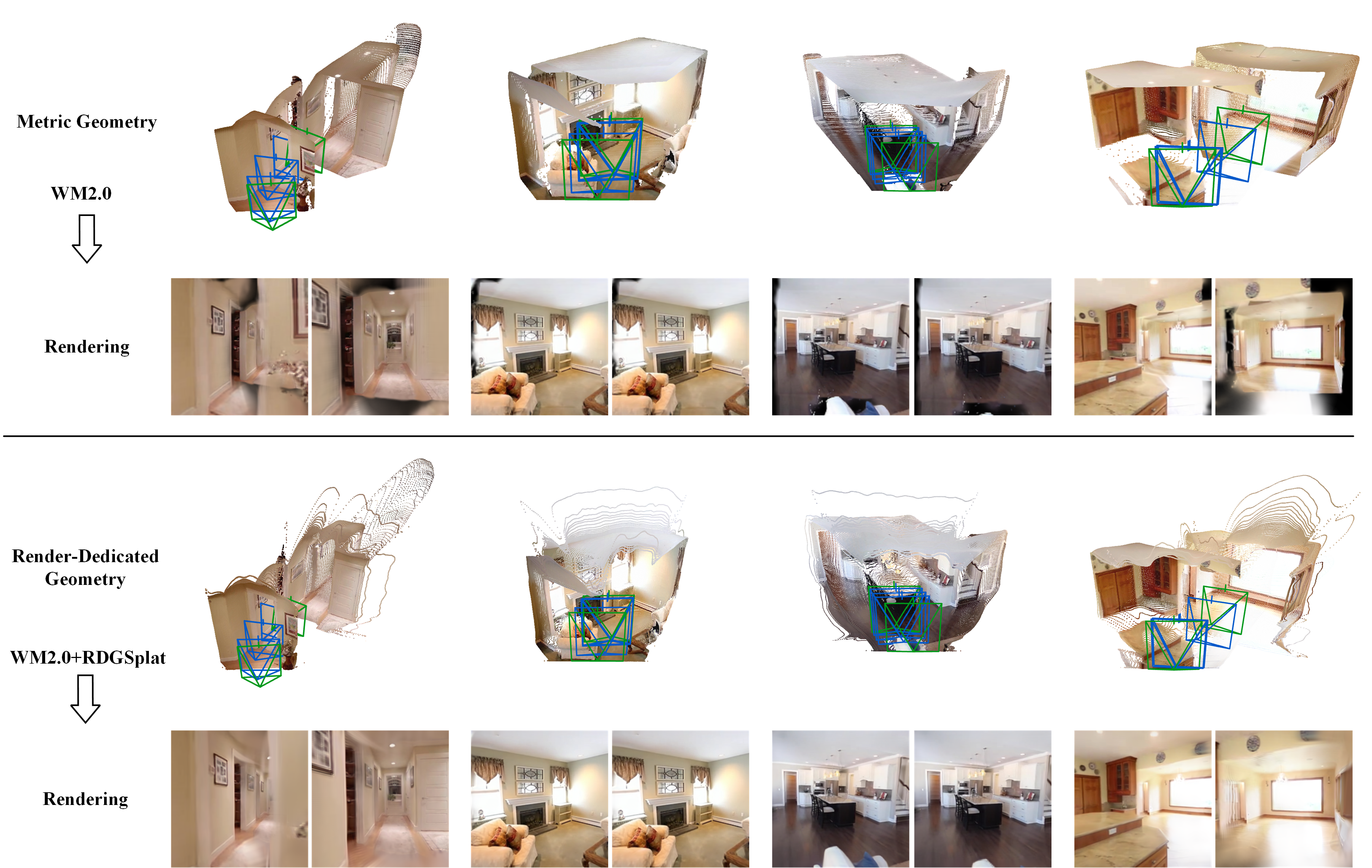}
  \caption{\textbf{The two geometries across scenes on RE10K.} Both are decoded from the same frozen backbone and the same context views, and the views rendered from each are shown beneath it. In all four scenes, the views rendered from the metric geometry leave the image borders empty, whereas those from the render-dedicated geometry cover the full frame.}
  \label{fig:geompair}
\end{figure*}

\begin{figure*}[htbp]
  \centering
  \includegraphics[width=0.9\linewidth]{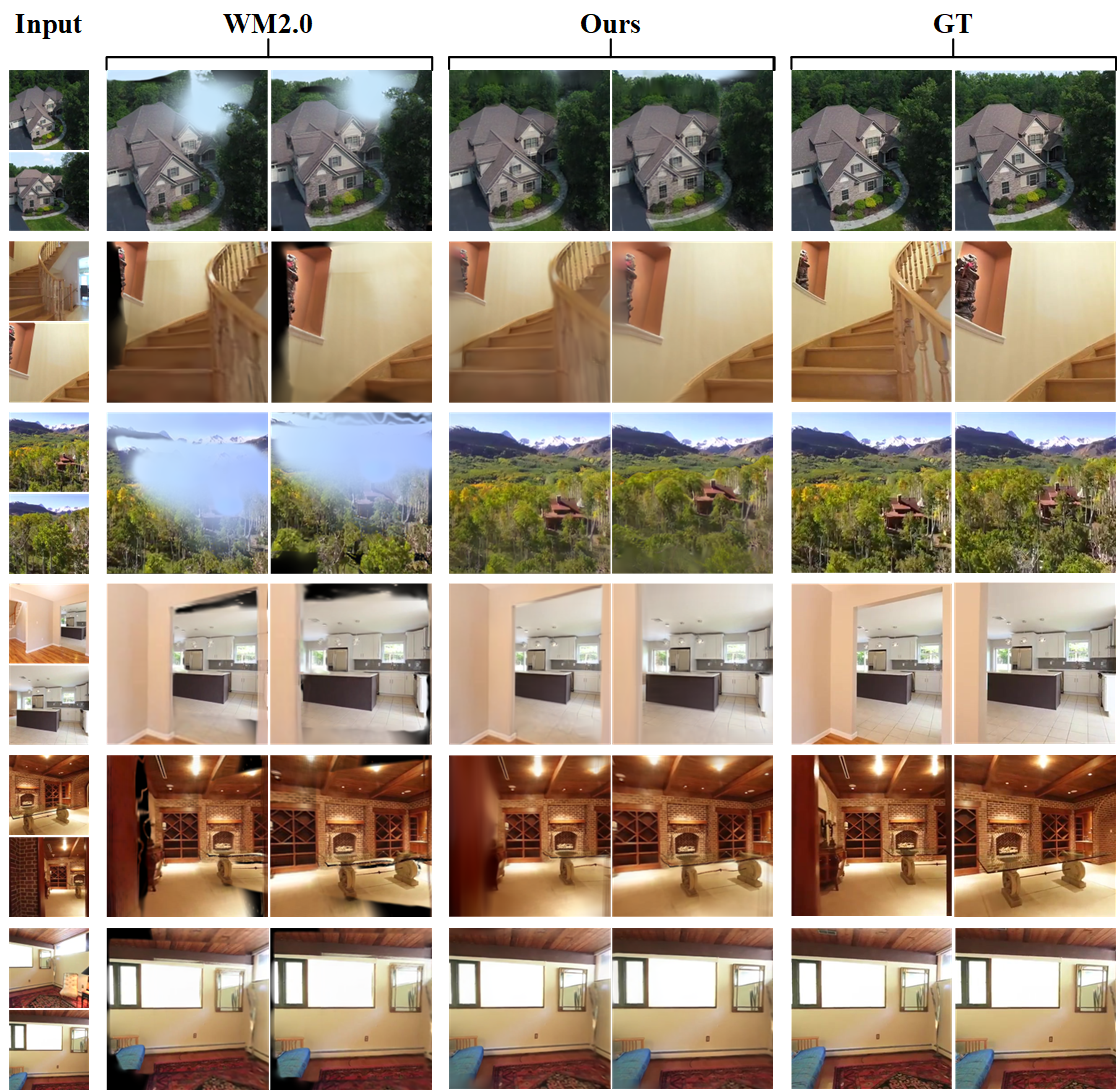}
  \caption{\textbf{Additional novel view synthesis on RE10K.} The leftmost column shows the two context views, and two held-out target views follow under each method.}
  \label{fig:morevis}
\end{figure*}

%%%%%%%%%%%%%%%%%%%%%%%%%%%%%%%%%%%%%%%%%%%%%%%%%%%%%%%%%%%%%%%%%%%%%%%%%%%%%%
%%%% B. Decomposition of RDGD
%%%%%%%%%%%%%%%%%%%%%%%%%%%%%%%%%%%%%%%%%%%%%%%%%%%%%%%%%%%%%%%%%%%%%%%%%%%%%%

\section{Decomposition of RDGD}
\label{sec:decomp}

RDGD carries three parts, which Table~3 of the main paper varies as one. Table~\ref{tab:decomp} adds them one at a time on RE10K~\cite{zhou2018stereo}, so that the 2.567\,dB reported there is read as three steps. Firstly, training the Gaussian decoder alone reaches 22.456\,dB, which recovers 1.538\,dB of the total and is the largest of the three. Then the duplicated depth head adds 0.882\,dB, which is where the primitives are free to leave the metric surface. Finally the residual pose head adds 0.147\,dB, the smallest step, and Row~4 reproduces the RDGD column of that table exactly. All three metrics improve at every step, and PSNR and SSIM keep this order.

Sec.~3.2 of the main paper argues that each duplicate should inherit the parameters of its source rather than start from noise, because both geometries describe the same observed surfaces. Table~\ref{tab:init} tests that argument by training the same three decoders from random weights, under the schedule and the data of Row~4. The inherited run reaches 23.485\,dB and the random one 17.449\,dB, a gap of 6.036\,dB. The random run also falls 3.469\,dB below the frozen backbone it was attached to. Within this schedule, the photometric objective alone therefore does not recover a geometry that renders better than the frozen one. This shows that the gain of the duplicates comes from the inherited parameters rather than from their capacity alone.

Table~\ref{tab:gsonly} compares fine-tuning the Gaussian decoder alone with RDGSplat, both trained for 200k iterations as in Table~1 of the main paper. The Gaussian decoder alone reaches 22.838\,dB, 1.920\,dB above the frozen backbone, whereas RDGSplat reaches 25.149\,dB. This indicates that the render depth and pose of RDGD, together with the representation adapted by TCA, further improve the rendering quality beyond what the decoder alone can attain.

\begin{table}[htbp]
\caption{\textbf{Decomposition of RDGD.} All rows are trained for 30k iterations, and Row~1 denotes WM2.0 with nothing attached. The best results are highlighted.}
\label{tab:decomp}
\centering
\setlength{\tabcolsep}{8pt}
\renewcommand{\arraystretch}{1.2}
\resizebox{\linewidth}{!}{
\begin{tabular}{lccc|ccc}
\toprule[1pt]
\multirow{2}{*}{} & \multirow{2}{*}{$g_s$} & \multirow{2}{*}{$g_d$} & \multirow{2}{*}{Res.\ pose}
& \multicolumn{3}{c}{\textbf{RE10K}} \\
\cmidrule(lr){5-7}
& & & & PSNR $\uparrow$ & SSIM $\uparrow$ & LPIPS $\downarrow$ \\
\midrule[0.6pt]
\midrule[0.6pt]
1 & --         & --         & --         & 20.918 & 0.747 & 0.210 \\
2 & \checkmark & --         & --         & 22.456 & 0.776 & 0.193 \\
3 & \checkmark & \checkmark & --         & 23.338 & 0.796 & 0.176 \\
4 & \checkmark & \checkmark & \checkmark & \textbf{23.485} & \textbf{0.801} & \textbf{0.175} \\
\bottomrule[1pt]
\end{tabular}
}
\end{table}

\begin{table}[htbp]
\caption{\textbf{Initialization of the duplicated decoders.} All rows are trained for 30k iterations, and Row~1 starts the three decoders from random weights. The best results are highlighted.}
\label{tab:init}
\centering
\setlength{\tabcolsep}{8pt}
\renewcommand{\arraystretch}{1.2}
\resizebox{0.8\linewidth}{!}{
\begin{tabular}{lc|ccc}
\toprule[1pt]
\multirow{2}{*}{} & \multirow{2}{*}{Inherited} & \multicolumn{3}{c}{\textbf{RE10K}} \\
\cmidrule(lr){3-5}
& & PSNR $\uparrow$ & SSIM $\uparrow$ & LPIPS $\downarrow$ \\
\midrule[0.6pt]
\midrule[0.6pt]
1 & --         & 17.449 & 0.545 & 0.407 \\
2 & \checkmark & \textbf{23.485} & \textbf{0.801} & \textbf{0.175} \\
\bottomrule[1pt]
\end{tabular}
}
\end{table}

\begin{table}[htbp]
\caption{\textbf{Fine-tuning the Gaussian decoder alone.} Both fine-tuned models are trained for 200k iterations, and RDGSplat corresponds to Table~1 of the main paper. The best results are highlighted.}
\label{tab:gsonly}
\centering
\setlength{\tabcolsep}{8pt}
\renewcommand{\arraystretch}{1.2}
\resizebox{0.8\linewidth}{!}{
\begin{tabular}{l|ccc}
\toprule[1pt]
\multirow{2}{*}{Method} & \multicolumn{3}{c}{\textbf{RE10K}} \\
\cmidrule(lr){2-4}
& PSNR $\uparrow$ & SSIM $\uparrow$ & LPIPS $\downarrow$ \\
\midrule[0.6pt]
\midrule[0.6pt]
WM2.0 & 20.918 & 0.747 & 0.210 \\
$g_s$ only & 22.838 & 0.783 & 0.185 \\
RDGSplat & \textbf{25.149} & \textbf{0.831} & \textbf{0.149} \\
\bottomrule[1pt]
\end{tabular}
}
\end{table}

%%%%%%%%%%%%%%%%%%%%%%%%%%%%%%%%%%%%%%%%%%%%%%%%%%%%%%%%%%%%%%%%%%%%%%%%%%%%%%
%%%% C. The Condition in TCA
%%%%%%%%%%%%%%%%%%%%%%%%%%%%%%%%%%%%%%%%%%%%%%%%%%%%%%%%%%%%%%%%%%%%%%%%%%%%%%

\section{The Condition in TCA}
\label{sec:cond}

TCA injects a block $B_l$ at each of eight sites and conditions each on $C$, or on $C_{\text{local}}$ within the patch encoder. Table~\ref{tab:cond} separates the two. Row~2 trains the same eight sites without any condition, so each carries its block $B_l$, a copy of layer $l$, and its scale $\gamma_l$. Row~3 removes only $C_{\text{prj}}$, so that every site receives $C_{\text{local}}$ and no term depends on the target view. The schedule, the data and the injection depths are those of the TCA-only model in Table~3 of the main paper. Adding the eight blocks alone reaches 21.520\,dB, which is 0.602\,dB above the frozen backbone. The target-independent terms $C_{\text{ray}}$ and $C_{\text{hf}}$ raise this to 22.213\,dB, and $C_{\text{prj}}$ adds a further 0.819\,dB, up to 23.032\,dB. This shows that most of what TCA contributes comes from the condition rather than from the capacity the blocks add. Within the condition, $C_{\text{prj}}$, the only term that requires the second evaluation to be repeated per target view, accounts for 0.819 of the 1.512\,dB.

\begin{table}[htbp]
\caption{\textbf{The condition in TCA.} All rows are trained for 30k iterations, and Row~1 denotes WM2.0 with nothing attached. The best results are highlighted.}
\label{tab:cond}
\centering
\setlength{\tabcolsep}{8pt}
\renewcommand{\arraystretch}{1.2}
\resizebox{\linewidth}{!}{
\begin{tabular}{lccc|ccc}
\toprule[1pt]
\multirow{2}{*}{} & \multirow{2}{*}{Blocks} & \multirow{2}{*}{$C_{\text{ray}}, C_{\text{hf}}$} & \multirow{2}{*}{$C_{\text{prj}}$}
& \multicolumn{3}{c}{\textbf{RE10K}} \\
\cmidrule(lr){5-7}
& & & & PSNR $\uparrow$ & SSIM $\uparrow$ & LPIPS $\downarrow$ \\
\midrule[0.6pt]
\midrule[0.6pt]
1 & --         & --         & --         & 20.918 & 0.747 & 0.210 \\
2 & \checkmark & --         & --         & 21.520 & 0.758 & 0.205 \\
3 & \checkmark & \checkmark & --         & 22.213 & 0.772 & 0.198 \\
4 & \checkmark & \checkmark & \checkmark & \textbf{23.032} & \textbf{0.788} & \textbf{0.186} \\
\bottomrule[1pt]
\end{tabular}
}
\end{table}

%%%%%%%%%%%%%%%%%%%%%%%%%%%%%%%%%%%%%%%%%%%%%%%%%%%%%%%%%%%%%%%%%%%%%%%%%%%%%%
%%%% D. Render-Dedicated Geometry on the Metric Benchmarks
%%%%%%%%%%%%%%%%%%%%%%%%%%%%%%%%%%%%%%%%%%%%%%%%%%%%%%%%%%%%%%%%%%%%%%%%%%%%%%

\section{Render-Dedicated Geometry on the Metric Benchmarks}
\label{sec:metric}

Table~\ref{tab:pose} compares the camera the pretrained head predicts against the camera the render branch uses. Both are scored together on the same 277 scenes, under the error definition of Table~2 of the main paper. Rotation is identical between the two, which Eq.~4 of the main paper guarantees by carrying $R_A$ over unaltered. Translation separates them: the median error rises from 1.589 to 1.609 degrees, and the AUC falls by 0.1 at 10 degrees while it is unchanged at 20. The render branch is therefore slightly less accurate as a camera estimate, as expected of a residual optimized for rendering rather than for camera accuracy. Depth admits no such comparison, because the second evaluation is defined only with respect to a target view, which the video depth benchmarks do not specify.

We further investigate the effect of fine-tuning the camera head directly under the photometric loss, in place of the residual pose head, while the rest of RDGSplat is unchanged. As shown in Table~\ref{tab:pose}, after 30k iterations its median translation error rises to 28.789 degrees, against 1.609 degrees for RDGSplat. Its rendering quality on RE10K also degrades to 14.693\,dB, 0.472 and 0.545 in PSNR, SSIM and LPIPS, against 25.149\,dB, 0.831 and 0.149 for RDGSplat. This demonstrates that the pretrained pose is necessary as an anchor, and that the residual pose head lets the render branch benefit from rendering supervision without losing this anchor.

Fig.~\ref{fig:geompair} extends the single scene of Fig.~2 of the main paper to four RE10K scenes. In each, the metric geometry forms dense, solid surfaces, whereas the render-dedicated geometry spreads them into thin layers, and only the views rendered from the latter cover the full frame.

\begin{table}[tbp]
\caption{\textbf{Two-view pose and rendering.} Median angular error in degrees, and AUC of the larger of the two errors, under the protocol of Table~2 of the main paper, with rendering measured on RE10K. The first two columns use the model of Table~1 of the main paper, where the metric geometry renders as the frozen WM2.0, and the last column fine-tunes the camera head directly in place of the residual pose head, trained for 30k iterations.}
\label{tab:pose}
\centering
\setlength{\tabcolsep}{8pt}
\renewcommand{\arraystretch}{1.2}
\resizebox{\linewidth}{!}{
\begin{tabular}{c|ccc}
\toprule[1pt]
& \textbf{\shortstack{Metric\\geometry}} & \textbf{\shortstack{Render-dedicated\\geometry}} & \textbf{\shortstack{Direct\\fine-tuning}} \\
\midrule[0.6pt]
\midrule[0.6pt]
Rot.\ median $\downarrow$   & 0.622 & 0.622 & 0.622 \\
Trans.\ median $\downarrow$ & 1.589 & 1.609 & 28.789 \\
AUC@10$^\circ$ $\uparrow$  & 71.6 & 71.5 & 6.4 \\
AUC@20$^\circ$ $\uparrow$  & 82.9 & 82.9 & 17.2 \\
\midrule[0.6pt]
PSNR $\uparrow$  & 20.918 & 25.149 & 14.693 \\
SSIM $\uparrow$  & 0.747 & 0.831 & 0.472 \\
LPIPS $\downarrow$ & 0.210 & 0.149 & 0.545 \\
\bottomrule[1pt]
\end{tabular}
}
\end{table}

%%%%%%%%%%%%%%%%%%%%%%%%%%%%%%%%%%%%%%%%%%%%%%%%%%%%%%%%%%%%%%%%%%%%%%%%%%%%%%
%%%% E. Inference Cost
%%%%%%%%%%%%%%%%%%%%%%%%%%%%%%%%%%%%%%%%%%%%%%%%%%%%%%%%%%%%%%%%%%%%%%%%%%%%%%

\section{Inference Cost}
\label{sec:cost}

As Table~\ref{tab:cost_supple} shows, RDGSplat adds latency to the frozen backbone but almost no memory. Beyond the pretrained heads, it computes a preliminary render depth and, for every target view, evaluates the backbone on the context views with TCA. Latency therefore rises from 262.3 to 688.5\,ms at two context views, and the ratio stays between 2.59 and 2.82 up to ten views, so the overhead grows no faster than the backbone itself. Peak memory, in contrast, rises by at most 0.06\,GB, since the two evaluations run in sequence and each frees its memory before the next.

\begin{table}[tbp]
\caption{\textbf{Inference cost.} One inference per row on a single NVIDIA A6000, with latency the median over eight scenes and three repetitions.}
\label{tab:cost_supple}
\centering
\setlength{\tabcolsep}{8pt}
\renewcommand{\arraystretch}{1.2}
\resizebox{\linewidth}{!}{
\begin{tabular}{cl|cc}
\toprule[1pt]
\textbf{Views} & \textbf{Method} & \textbf{Latency (ms)} $\downarrow$ & \textbf{Peak memory (GB)} $\downarrow$ \\
\midrule[0.6pt]
\midrule[0.6pt]
\multirow{2}{*}{2}  & WM2.0~\cite{team2026hy} & 262.3 & 7.74 \\
                    & WM2.0 + RDGSplat & 688.5 & 7.80 \\
\midrule[0.6pt]
\multirow{2}{*}{3}  & WM2.0 & 286.9 & 7.85 \\
                    & WM2.0 + RDGSplat & 761.0 & 7.90 \\
\midrule[0.6pt]
\multirow{2}{*}{5}  & WM2.0 & 351.4 & 8.11 \\
                    & WM2.0 + RDGSplat & 990.3 & 8.11 \\
\midrule[0.6pt]
\multirow{2}{*}{10} & WM2.0 & 526.7 & 8.71 \\
                    & WM2.0 + RDGSplat & 1361.8 & 8.72 \\
\bottomrule[1pt]
\end{tabular}
}
\end{table}

%%%%%%%%%%%%%%%%%%%%%%%%%%%%%%%%%%%%%%%%%%%%%%%%%%%%%%%%%%%%%%%%%%%%%%%%%%%%%%
%%%% F. Additional Qualitative Results
%%%%%%%%%%%%%%%%%%%%%%%%%%%%%%%%%%%%%%%%%%%%%%%%%%%%%%%%%%%%%%%%%%%%%%%%%%%%%%

\section{Additional Qualitative Results}
\label{sec:qual}

Further qualitative comparisons are provided in Fig.~\ref{fig:morevis}, which extends Fig.~6 of the main paper to six RE10K scenes. Across these scenes, the frozen backbone covers distant regions such as the sky with blurry blobs, and leaves empty or blurred regions near the image borders and occlusion boundaries. RDGSplat renders the same views largely without these artifacts.

\end{document}